\documentclass{article}

\usepackage{PRIMEarxiv}

\usepackage[utf8]{inputenc} % allow utf-8 input
\usepackage[T1]{fontenc}    % use 8-bit T1 fonts
\usepackage{hyperref}       % hyperlinks
\usepackage{url}            % simple URL typesetting
\usepackage{booktabs}       % professional-quality tables
\usepackage{amsfonts}       % blackboard math symbols
\usepackage{nicefrac}       % compact symbols for 1/2, etc.
\usepackage{microtype}      % microtypography
\usepackage{lipsum}
\usepackage{fancyhdr}       % header
\usepackage{listings}
\usepackage{xcolor}
\usepackage{hyperref}
\usepackage{url}
\usepackage{graphicx}
\usepackage{fancyvrb}
\usepackage{fvextra}
\usepackage{algorithm}
\usepackage{algpseudocode}
\usepackage{amsmath}

\title{ToolBridge: An Open-Source Dataset to Equip LLMs with External Tool Capabilities}

\author{
  Zhenchao Jin \textsuperscript{1}, Mengchen Liu \textsuperscript{2}, Dongdong Chen \textsuperscript{2}, Lingting Zhu \textsuperscript{1}, Yunsheng Li \textsuperscript{2}, Lequan Yu \textsuperscript{1}\\
  \textsuperscript{1} The University of Hong Kong, \textsuperscript{2} Microsoft \\
  \texttt{\{blwx96@connect., ltzhu99@connect., lqyu@\}hku.hk} \\
  \texttt{\{mengcliu, dongdong.chen, yunshengli\}@microsoft.com} \\
}

\begin{document}
\maketitle

\begin{abstract}
  Through the integration of external tools, large language models (LLMs) such as GPT-4o and Llama 3.1 significantly expand their functional capabilities, evolving from elementary conversational agents to general-purpose assistants. 
  We argue that the primary drivers of these advancements are the quality and diversity of the training data.
  However, the existing LLMs with external tool integration provide only limited transparency regarding their datasets and data collection methods, which has led to the initiation of this research.
  Specifically, in this paper, our objective is to elucidate the detailed process involved in constructing datasets that empower LLMs to effectively learn how to utilize external tools and make this information available to the public through the introduction of ToolBridge.
  ToolBridge proposes to employ a collection of general open-access datasets as its raw dataset pool and applies a series of strategies to identify appropriate data entries from the pool for external tool API insertions.
  By supervised fine-tuning on these curated data entries, LLMs can invoke external tools in appropriate contexts to boost their predictive accuracy, particularly for basic functions including data processing, numerical computation, and factual retrieval.
  Our experiments rigorously isolates model architectures and training configurations, focusing exclusively on the role of data.
  The experimental results indicate that LLMs trained on ToolBridge demonstrate consistent performance improvements on both standard benchmarks and custom evaluation datasets. 
  All the associated code and data will be open-source at \href{https://github.com/CharlesPikachu/ToolBridge}{https://github.com/CharlesPikachu/ToolBridge}, promoting transparency and facilitating the broader community to explore approaches for equipping LLMs with external tools capabilities.
\end{abstract}

\section{Introduction}
Large language models (LLMs) have revolutionized natural language processing, excelling in tasks including question answering, summarization, and text generation \cite{jiang2023mistral, gunter2024apple, team2024gemma}.
Despite the impressive achievements of LLMs, they persistently underperform in fundamental areas, such as arithmetic or factual lookup, where introducing external tools like calculators and search engines can effectively provide solutions \cite{cobbe2021training,thoppilan2022lamda,schick2024toolformer,shen2024hugginggpt,gao2023pal,achiam2023gpt,zhao2024let}.

Consequently, there has been a rise in research efforts aimed at equipping LLMs with the ability to leverage external tools. 
For example, Llama 3.1 \cite{dubey2024llama} is trained to utilize Brave Search, Python interpreter and Wolfram Alpha API by using synthetic data combined with human-annotated data. 
Toolformer \cite{schick2024toolformer} adopts a self-supervised training strategy to equip LLMs with the capability to leverage diverse tools effectively.
However, these existing LLMs tend to either only open-source their trained models and code or restrict access to the models through usage interfaces \cite{achiam2023gpt, dubey2024llama, gunter2024apple}.
The data underlying the training of these models is commonly withheld from public disclosure, enveloped in secrecy.

The widespread availability of pre-trained models stands in stark contrast to the limited transparency surrounding their training datasets, which impedes the community’s progress in advancing discoveries in this area.
This emphasizes the critical need to explore ways to create high-quality, publicly available benchmark datasets for training LLMs to utilize external tools to facilitate their reasoning process.
  
\begin{figure}[t]
\centering
\includegraphics[width=1.0 \textwidth]{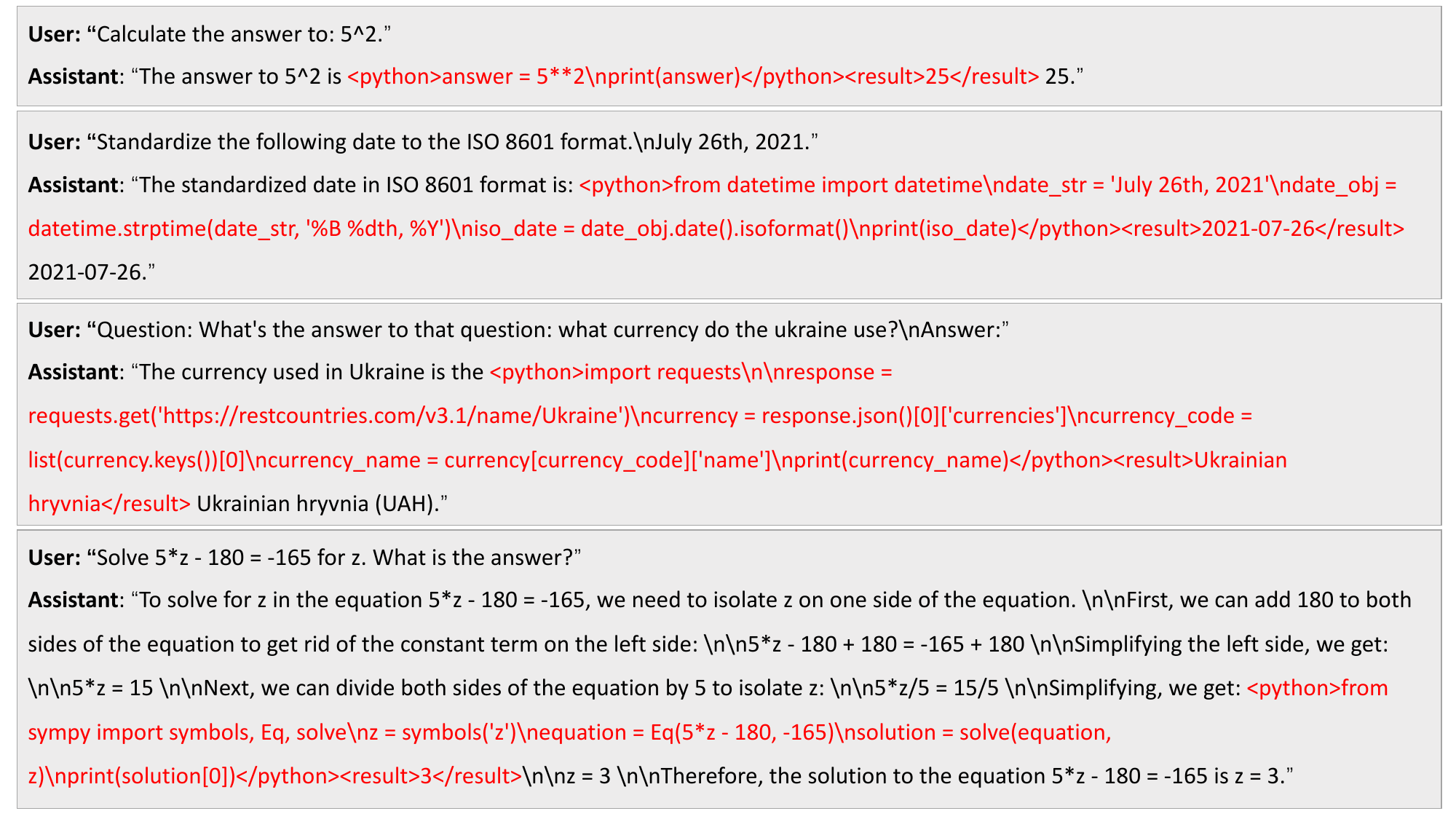} \label{fig.datasamples}
\caption{\textbf{Examples of data entries from ToolBridge.} The proposed pipeline is capable of generating a large volume of data entries to facilitate model learning to utilize external tools. }
\vspace{-0.3cm}
\end{figure}

In response to the challenges discussed, this paper proposes a pipeline for the large-scale creation of datasets designed to train LLMs in leveraging external tools. 
Specifically, we begin by aggregating a substantial collection of open-source datasets used for LLM supervised fine-tuning (SFT) from the community, which circumvents proprietary concerns (\emph{e.g.}, copyright issues).
Upon establishing this dataset pool, we propose a systematic approach to assist in the identification of valuable data entries and convert them into a standardized format.
Finally, valuable data entries filtering is conducted to further boost the quality of these converted data entries and ToolBridge is constructed.
As demonstrated in Figure \ref{fig.datasamples}, we showcase the data entries from ToolBridge, which are intended to guide LLMs in understanding how to incorporate external tools in appropriate contexts, thereby improving the accuracy and reliability of their outputs.
For instance, the third example in Figure \ref{fig.datasamples} serves to instruct LLMs in leveraging the \emph{requests} module to gather factual information from web sources. 
Moreover, LLMs can utilize the fourth example to understand how \emph{sympy} library can be applied to solve linear equation of one variable.

In summary, the contributions of this paper are as follows,

\begin{itemize}
    \item We propose a pipeline capable of producing large volumes of entries for training LLMs to incorporate various external tools.
    The collection of over 178,000 yielded data entries, named ToolBridge, will be open-sourced to the community, marking a significant advancement in the transparency and accessibility of the data for training the large language models to leverage external tools.
    
    \item Our experimental results indicate that LLMs supervised fine-tuned on our ToolBridge can achieve consistent performance improvements on several standard benchmarks.
    
    \item We also propose to curate a set of data entries to examine the abilities of LLMs in numerical calculation, data processing and factual retrieval before and after supervised fine-tuning on ToolBridge.
    Experimental outcomes reaffirm the effectiveness of ToolBridge.
\end{itemize}

This is, to our knowledge, the first research in the domain of equipping LLMs with external tools that open-sources the training data.
We anticipate that ToolBridge will enable the community to further investigate the ability of LLMs to use external tools, thereby advancing LLMs from basic conversational models to versatile general-purpose assistants.

\section{Related Work}

\noindent \textbf{Tool Use for LLMs.} Enabling LLMs to use tools like search engines and code interpreters significantly broadens the range of tasks they can address and improves their accuracy in tasks involving data processing, numerical computation and factual retrieval \cite{nakano2021webgpt,schick2022peer,gao2023pal,achiam2023gpt,dubey2024llama,schick2024toolformer}.
For instance, Komeili et al. \cite{komeili2021internet} proposed to enable the models to use a search engine by learning to generate an internet search query based on the context, and then condition its final response on the search results.
Cobbe et al. \cite{cobbe2021training} facilitated LLMs' usage of a calculator during testing by training LLMs with calculation annotations injected into the training set. 
Thoppilan et al. \cite{thoppilan2022lamda} assisted LLMs in invoking external tools from a toolset, comprising an information retrieval system, a calculator, and a translator, by training them to produce a special string \emph{TS}.
Gao et al. \cite{gao2023pal} proposed adopting LLMs to interpret natural language problems and generate programs as intermediate reasoning steps, while delegating the solution process to a runtime environment like a Python interpreter.
HuggingGPT \cite{shen2024hugginggpt} demonstrated how LLMs can act as controllers to manage and integrate various AI models across different domains and modalities.
Toolformer \cite{schick2024toolformer} allowed LLMs to learn how to employ external tools through a self-supervised learning approach.
Of late, GPT-4o \cite{achiam2023gpt}, Llama 3.1 \cite{dubey2024llama} and Apple LLMs \cite{gunter2024apple} and several other works \cite{zhao2024let} further strengthened LLMs' ability to use external tools through improvements in models, training data, and other areas.

Although these works are highly commendable, they rarely open-source the data required to train their models, which is essential for the community to further research and refine their contributions.
This paper presents a pipeline targeted at yielding data entries for training models in external tool utilization, along with open-sourcing all data entries produced using this methodology.
This open access facilitates the development of more effective and efficient algorithms for the next generation of LLMs integrated with external tool functionalities.

\begin{figure}[t]
\centering
\includegraphics[width=0.95\textwidth]{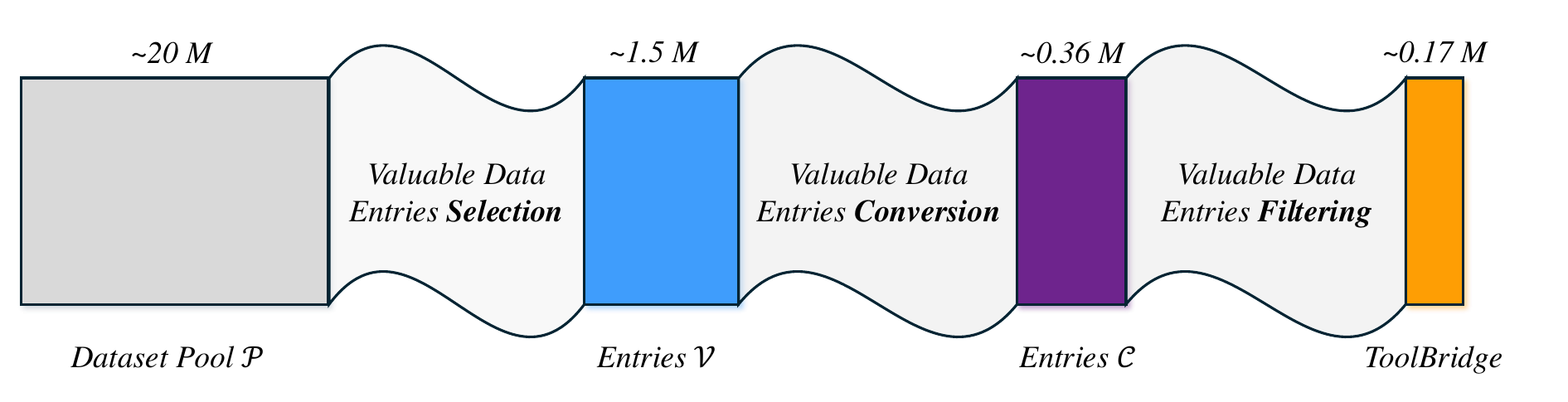} \label{fig.overview}
\vspace{-0.3cm}
\caption{\textbf{Overview of the construction pipeline for ToolBridge.}  The construction of ToolBridge follows three main steps: identifying valuable data entries from the dataset pool, converting these valuable data entries to standardized format, and finally conducting a thorough cleanup of the converted data entries.}
\vspace{-0.3cm}
\end{figure}

\noindent \textbf{Tool Use Benchmark Datasets.}
Previous research on benchmark datasets associate with tool use primarily focused on evaluating the ability of models to utilize external tools \cite{srinivasan2023nexusraven,zhuang2023toolqa, patil2023gorilla,xu2023tool,qin2023toolllm}.
Specifically, the datasets proposed in Kasai et al. \cite{kasai2024realtime} and Chen et al. \cite{chen2021dataset} were used to investigate how text retrievers improve the performance of LLMs on timely information.
ToolQA \cite{zhuang2023toolqa} proposed to develop tool-based question-answer pairs to assess LLMs' capability in arriving at accurate solutions. 
Huang et al. \cite{huang2023metatool} suggested setting various tasks to evaluate the proficiency of LLMs in tool usage awareness and the ability to select appropriate tools.
Xu et al. \cite{xu2023tool} developed a benchmark focused on tool manipulation, featuring a range of software tools applicable to real-world scenarios.
Given the scarcity of open-source large-scale datasets for equipping LLMs with external tool usage capabilities, we present ToolBridge which is a dataset of more than 178,000 data entries to support LLMs in effectively learning how to utilize external tools.

\section{ToolBridge}

Previous models like GPT-4o and Llama 3.1 only provide limited information on how they curate the data to train LLMs to leverage external tools. 
To address the lack of transparency in training data, we propose a generic pipeline for constructing large-scale datasets from public sources to enable LLMs to use external tools.
As illustrated in Figure~\ref{fig.overview}, the whole pipeline follows three main steps: valuable data entries \textbf{selection}, \textbf{conversion} and \textbf{filtering}.

\begin{table}[t]
    \centering
    \resizebox{0.75\textwidth}{!}{%
    \begin{tabular}{c|c|c|c}
        \hline
        \textbf{Source} & \textbf{\# of Entries} & \textbf{Source} & \textbf{\# of Entries} \\ \hline
        School Math 0.25M  & 248,481 & LIMA & 1,330 \\ \hline
        code\_instructions\_120k\_alpaca & 121,959 & TigerBot   & 1,199,030 \\ \hline
        Platypus & 24,926 & TSI-v0 & 5,607,620 \\ \hline
        ShareGPT90K & 90,665 & LaMini-Instruction & 2,585,615 \\ \hline
        WizardLM\_Orca &54,974  & Bactrian-X & 67,017 \\ \hline
        WizardLM\_evol\_instruct\_70k & 70,000  & Baize & 210,311 \\ \hline
        tiny-codes & 1,632,309 & COIG & 178,246 \\ \hline
        WizardLM evolve\_instruct V2 & 143,000  & MOSS SFT  & 1,074,551 \\ \hline
        No Robots & 10,000 & AlpacaDataCleaned  & 51,760 \\ \hline
        ign\_clean\_instruct\_dataset\_500k & 508,620  & GPT-4all  & 808,812 \\ \hline
        GPT-4-LLM & 113,003 & Alpaca  & 52,002 \\ \hline
        ChatAlpaca & 20,000  & self-instruct  & 82,439 \\ \hline
        OpenOrca & 4,233,923  &   &  \\ \hline
    \end{tabular}}
    \caption{The composition of our dataset pool $\mathcal{P}$ to construct ToolBridge.}
    \label{tab.datapool}
    \vspace{-0.5cm}
\end{table}

\subsection{Dataset Pool Construction}

Our research starts with a comprehensive review of the data accessible in the community for the purpose of supervised fine-tuning (SFT).
Table \ref{tab.datapool} summarizes the results.
Owing to the diverse range of teams contributing the SFT datasets, there is significant heterogeneity in their formats, which poses challenges for effective model training.
Hence, we first reformat all candidate datasets into a standardized ChatML format for further processing:

\begin{center}
\begin{BVerbatim}
entry = [{"role": "user", "content": "..."}, 
         {"role": "assistant", "content": "..."}, 
         ...].
\end{BVerbatim}
\end{center}

After reformatting all the datasets, we construct the dataset pool as $\mathcal{P}=\{(\mathcal{D}_i, W_i, Q_i) ~|~ i \ge 0 \}$, where $\mathcal{D}_i$ denotes one candidate dataset, 
$W_i$ measures the proportion of valuable entries for each candidate dataset, and $Q_i$ serves as a metric for assessing the data quality of each candidate dataset.

Practically, to obtain $W_i$, we first perform random sampling on $\mathcal{D}_i$ to produce a subset $\mathcal{S}_i$.
Llama3 \cite{touvron2023llama} is then applied to judge the appropriateness of each data entry in $\mathcal{S}_i$ for external tool invocation to enhance reasoning,
where the prompt employed is shown in Appendix \ref{sec.llama3prompts}.
In generally, if a data entry is deemed suitable for invoking external tools to help LLMs' reasoning process, we label it as a valuable data entry.
At last, we determine $W_i$ as the ratio between the number of valuable data entries and the total number of data entries in $\mathcal{S}_i$.

Additionally, we observe that some candidate datasets within the dataset pool are partially sourced from the Internet via web scraping,
resulting in the presence of meaningless HTML tags and other irrelevant content.
Thus, we incorporate $Q_i$ as an additional metric to evaluate the data quality of each candidate dataset.
To compute $Q_i$, we randomly sample $N$ data entries from $\mathcal{D}_i$ and conduct a manual review to identify any presence of irrelevant characters or content.
$Q_i$ is then derived as the fraction of data entries devoid of irrelevant content over the total number of sampled entries $N$.

\subsection{Valuable Data Entries Selection} \label{sec.vdes}

Upon constructing the dataset pool $\mathcal{P}$, due to the large scale of candidate data entries, we propose to select 10 million data entries from $\mathcal{P}$ for further processing according to the dataset attributes $W_i$ and $Q_i$.
Particularly, we first arrange $\mathcal{D}_i$ in descending order based on the value of $Q_i \times W_i$.
Then, the data entries are selected from the top-ranked datasets sequentially until the overall volume reaches 10 million entries.

Subsequently, Llama3-70B with the prompt detailed in Appendix \ref{sec.llama3prompts} is applied to ascertain whether each entry within the 10 million samples is appropriate for LLMs to enhance reasoning via introducing external tools.
And we represent the collection of these appropriate data entries with $\mathcal{V}$ (\emph{i.e.}, valuable data entries).
Table \ref{tab.valuabledata} presents the distribution of $\mathcal{V}$ across the respective datasets they originate from.
It is observed that leveraging the capabilities of Llama3-70B, we can refine the 10 million data entries down to 1,527,153 valuable entries.

\begin{table}[h]
    \centering
    \resizebox{0.8\textwidth}{!}{%
    \begin{tabular}{c|c|c|c}
        \hline
        \textbf{Source} & \textbf{\# of Entries} & \textbf{Source} & \textbf{\# of Entries} \\ \hline
        School Math 0.25M  & 205,996 & ChatAlpaca & 2,643 \\ \hline
        Platypus  & 7,776 & ShareGPT90K & 24,348 \\ \hline
        WizardLM\_Orca  & 8,659 & WizardLM\_evol\_instruct\_70k & 28,293 \\
        \hline
        WizardLM evolve\_instruct V2 & 5,399 & MOSS SFT & 136,603 \\ \hline
        TigerBot & 182,249 & GPT-4all & 47,627 \\ \hline
        COIG & 15,181 & LIMA & 309 \\ \hline
        AlpacaDataCleaned & 13,805 & GPT-4-LLM & 9,978 \\ \hline
        Bactrian-X & 3,313 & OpenOrca & 834,974 \\ \hline
    \end{tabular}}
    \caption{The composition of the selected valuable data entries $\mathcal{V}$.}
    \label{tab.valuabledata}
    \vspace{-0.5cm}
\end{table}

\subsection{Valuable Data Entries Conversion}

Following valuable data entries selection, we further convert the selected data entries, allowing the language models to learn how to invoke external tools effectively within the proper context to support their reasoning process.
In particular, we draw on previous approaches~\cite{schick2024toolformer,dubey2024llama,thoppilan2022lamda} by embedding special characters in each selected entry to enable the external tool invocation, 
and LLMs are employed to pinpoint the appropriate context for calling external tool APIs.

As illustrated in the examples in Figure \ref{fig.datasamples}, \verb|<python>| and \verb|</python>| are represented as a pair of special tokens.
While the content enclosed by the special tokens specifies the Python invocation for external tool use.
During the construction of ToolBridge, we predominantly utilize GPT-4o-mini to insert the special tokens in the correct context within each data entry identified in Section \ref{sec.vdes}, as well as to create the associated code for invoking the tools.
To facilitate the return of tool execution results, we examine the code format utilized by GPT-4o when calling Python API as part of its reasoning process while answering user queries.
It is observed that the final result is always printed at the end of the code.
In line with the approach of GPT-4o, we include a directive in the prompt for GPT-4o-mini to print the final result of the tool execution as the concluding line of the code (refer to Appendix \ref{sec.gpt40miniprompts}). 

Following GPT-4o-mini's processing of $\mathcal{V}$, we extract the code segments between \verb|<python>| and \verb|</python>|, execute them, and finally insert the captured output after corresponding \verb|</python>|,
where the output will be wrapped within another pair of special tokens, denoted as \verb|<result>| and \verb|</result>|.
To summarize, the tool invocation in ToolBridge can be formatted as,

\begin{center}
\begin{BVerbatim}
<python>generated code</python><result>captured output</result>.
\end{BVerbatim}
\end{center}

During the reasoning process of the models, it is sufficient to check for the presence of the pre-defined special tokens \verb|<python>| and \verb|</python>| during generation.
If identified, Python interpreter will be adopted to execute the enclosed code between the special tokens, and the executed result is wrapped in \verb|<result>| and \verb|</result>| to condition further text generation.
Algorithm \ref{alg.processst} describes the primary inference process of LLMs post-SFT on the ToolBridge dataset.

\begin{algorithm} [h]
\caption{Process Special Tokens During Inference} \label{alg.processst}
\begin{algorithmic}[1]
\Require Python interpreter and $inputs$.
\Ensure Execute the code enclosed by \texttt{<python>} and \texttt{</python>}, and use the captured output as a condition for the subsequent text generation.

\State Initialize an empty list $outputs$
\State Initialize $condition \gets \text{None}$
\While{$outputs$ is empty \textbf{or} $outputs[-1] \neq \texttt{<|end\_of\_text|>}$}
    \State $output \gets \text{LLM}(inputs)$
    \If {$output = \texttt{<python>}$}
        \State $start \gets \text{length of } outputs$ + \text{length of \texttt{<python>}}
    \ElsIf {$output = \texttt{</python>}$}
        \State Extract substring $code \gets outputs[start:]$
        \State $condition \gets \text{ExecutePython}(code)$
    \EndIf
    \State Append $output$ to both $inputs$ and $outputs$
    \If {$condition$ is not \texttt{None}}
        \State Append $condition$ to both $inputs$ and $outputs$
        \State Reset $condition \gets \text{None}$
    \EndIf
\EndWhile
\State Post-process and return $outputs$
\end{algorithmic}
\end{algorithm}

Table \ref{tab.convertedvaluabledata} summarizes the composition of the data entries after converting $\mathcal{V}$, denoted as $\mathcal{C}$.

\begin{table}[h]
    \centering
    \resizebox{0.8\textwidth}{!}{%
    \begin{tabular}{c|c|c|c}
        \hline
        \textbf{Source} & \textbf{\# of Entries} & \textbf{Source} & \textbf{\# of Entries} \\ \hline
        School Math 0.25M  & 150,104 & ChatAlpaca & 116 \\ \hline
        Platypus  & 82 & ShareGPT90K & 3 \\ \hline
        WizardLM\_Orca  & 88 & WizardLM\_evol\_instruct\_70k & 3,716 \\
        \hline
        WizardLM evolve\_instruct V2 & 8 & OpenOrca & 101,715 \\ \hline
        TigerBot & 66,793 & GPT-4all & 8,936 \\ \hline
        COIG & 7,877 & LIMA & 107 \\ \hline
        AlpacaDataCleaned & 4,510 & GPT-4-LLM & 2,577 \\ \hline
        Bactrian-X & 84 &  &  \\ \hline
    \end{tabular}}
    \caption{The composition of the converted data entries $\mathcal{C}$.}
    \label{tab.convertedvaluabledata}
\end{table}

It is worth noting that the data entries in $\mathcal{C}$, totaling 364,605, are considerably fewer compared to those in $\mathcal{V}$ (1,527,153). 
We attribute this to the following factors,

\begin{itemize}
    \item The returned data entries that lack Python code inserted by GPT-4o-mini are eliminated. 
    Approximately 19.2\% of the data entries in $\mathcal{V}$ fall under this category, which indicates that GPT-4o-mini considers these data entries as not requiring additional tools to aid in LLMs' reasoning.
    
    \item The returned data format did not match the expected structure, resulting in parsing failure. 
    Instances include an unequal count of \texttt{<python>} and \texttt{</python>} tokens, modifications to the original content alongside Python code insertion, to name a few.
    Such data entries account for approximately 27.2\% of $\mathcal{V}$.
    
    \item The request to GPT-4o-mini fails. These data entries constitute approximately 2.1\% of $\mathcal{V}$.

    \item With Python's \emph{ast} library, we filter out the returned data entries where the inserted code is only an assignment followed by a print statement. 
    Such data entries represent about 4.8\% of $\mathcal{V}$. 
    We provide the detailed information about the algorithm we adopt to filter such data entries in Appendix \ref{sec.identifyingcodepatterns}.
    
    \item We filter around 22.8\% entries of $\mathcal{V}$ to remove tool calls that could not be executed or the execution time of the tool exceeded 30 seconds (refer to Appendix \ref{sec.removebadcode} for details).
\end{itemize}

In summary, 23.9\% of the entries from $\mathcal{V}$ remain in the converted data entries $\mathcal{C}$.

\subsection{Data Entries Filtering by Consistency Validation}

In practice, we observe that LLMs trained on $\mathcal{C}$ do not always base their subsequent content on the results produced by the yielded Python code during inference.
So, we conduct a re-evaluation of the data entries within $\mathcal{C}$ and observe that the execution results from the code generated by GPT-4o-mini also does not always align with the ensuing text, which explains LLMs' sporadic inconsistencies between tool execution results and further contents during inference.

To alleviate the issues above, we propose to filter the data entries in $\mathcal{C}$ where the tool execution results are inconsistent with the following text,
which is achieved by checking if the execution results are included in the subsequent text in our implementation.
Upon completing filtering, the open-source dataset ToolBridge is finally constructed. 
In Appendix \ref{sec.compareconsistentoutputs}, we compare the generated text of Llama3-8B after SFT on $\mathcal{C}$ and ToolBridge, which demonstrates the necessity for the data entries filtering by consistency validation.

The data sources that make up ToolBridge, totaling 178,023 entries, are outlined in Table \ref{tab.toolbridge}, which represents 48.8\% of the total data entries in $\mathcal{C}$.

\begin{table}[h]
    \centering
    \resizebox{0.8\textwidth}{!}{%
    \begin{tabular}{c|c|c|c}
        \hline
        \textbf{Source} & \textbf{\# of Entries} & \textbf{Source} & \textbf{\# of Entries} \\ \hline
        School Math 0.25M  & 100,836 & ChatAlpaca & 17 \\ \hline
        Platypus  & 35 & ShareGPT90K & 3 \\ \hline
        WizardLM\_Orca  & 29 & WizardLM\_evol\_instruct\_70k & 794 \\
        \hline
        WizardLM evolve\_instruct V2 & 1 & OpenOrca & 46,449 \\ \hline
        TigerBot & 22,306 & GPT-4all & 2,616 \\ \hline
        COIG & 2,706 & LIMA & 27 \\ \hline
        AlpacaDataCleaned & 1,129 & GPT-4-LLM & 1,043 \\ \hline
        Bactrian-X & 32 &  &  \\ \hline
    \end{tabular}}
    \caption{The final composition of our proposed ToolBridge.}
    \label{tab.toolbridge}
\end{table}

\section{EXPERIMENTS}

In this section, we first present the statistics of ToolBridge and then investigate whether LLMs with SFT on ToolBridge can use external tools to facilitate their reasoning process.
The validation includes three parts: 
(1) by comparing LLMs' performance on standard benchmarks before and after SFT on ToolBridge (Section \ref{sec.rsb}); 
(2) by evaluating the accuracy of the models on the custom datasets (Section \ref{sec.rcb});
(3) by comparing the qualitative results of LLMs (Section \ref{sec.qr}).

\subsection{Dataset Statistics of ToolBridge}

In Table \ref{tab.toolbridge}, we have exhibited 15 source datasets involved in the data entries of ToolBridge, along with their respective composition ratios.
To prevent any confusion for these datasets (\emph{e.g.}, other datasets with the same name and the same datasets with different versions), we also provide download links for these datasets in Appendix \ref{sec.datasourcelink}.

Besides, we provide a summary of the frequency of external tool calls in ToolBridge and the variety of Python packages used for these calls, as presented in Table \ref{tab.apistatistics}. 
By comparing Table \ref{tab.toolbridge} and Table \ref{tab.apistatistics}, we can observe that the majority of the data entries in ToolBridge involve only a single external tool call.
Moreover, from Table \ref{tab.apistatistics}, we can observe that the data entries in ToolBridge originating from datasets with narrower topics (\emph{e.g.}, School Math 0.25M) generally utilize fewer kinds of Python packages compared to those from datasets with broader topics (\emph{e.g.}, TigerBot), which is consistent with logic.
To summarize, there are 183,147 external tool calls in ToolBridge, using a total of 62 Python packages, including \emph{requests}, \emph{math}, \emph{datetime}, \emph{sklearn}, to name a few.

\begin{table}[H]
    \centering
    \resizebox{1.0\textwidth}{!}{%
    \begin{tabular}{c|c|c|c|c|c}
        \hline
        \textbf{Source} & \textbf{\# of Tool Use} & \textbf{\# of Libraries} & \textbf{Source} & \textbf{\# of Tool Use} & \textbf{\# of Libraries} \\ \hline
        School Math 0.25M  &104,983 & 8 & ChatAlpaca &73 & 1 \\ \hline
        Platypus  &36 & 3 & ShareGPT90K &3 & 0 \\ \hline
        WizardLM\_Orca  &33 &4 & WizardLM\_evol\_instruct\_70k &836 & 22 \\
        \hline
        WizardLM evolve\_instruct V2 &1 &0 & OpenOrca &46,832 & 28 \\ \hline
        TigerBot &22,507 & 40 & GPT-4all &2,870 & 42 \\ \hline
        COIG &2,719 & 28 & LIMA &27 & 2 \\ \hline
        AlpacaDataCleaned &1,139 & 23 & GPT-4-LLM &1,052 & 14 \\ \hline
        Bactrian-X &36 & 3 &  &  \\ \hline
    \end{tabular}}
    \caption{Statistics of the usage of external tools in ToolBridge.}
    \label{tab.apistatistics}
    \vspace{-0.5cm}
\end{table}

\subsection{Experimental Setup}

\noindent \textbf{Baseline Models.} 
Our experiments involve four baseline models: the base model of Llama2-7B and Llama3-8B.
Also, we remove all the external tool invocation sections in each data entry in ToolBridge (denote as ToolBridge$^{\S}$) and report the results of Llama2-7B and Llma3-8B SFT on ToolBridge$^{\S}$ as two additional baseline models.

\noindent \textbf{Benchmark Datasets.}
The standard benchmark datasets used in our experiments include GSM 8k \cite{cobbe2021training}, GSM Plus \cite{li2024gsm}, MathBench \cite{liu2024mathbench} and Stanford WebQA \cite{berant-etal-2013-semantic}.
For GSM 8k and GSM Plus, we conduct performance evaluation on their respective test sets under few-shot setting, where we leverage a fixed CoT-n-shot prompt template, as outlined in \cite{li2024gsm}.
For MathBench, we report results on MathBench-A, where we transform the multiple-choice questions in the College, High and Middle categories into a question-and-answer format for CoT-n-shot evaluation.
To differentiate from original MathBench, we refer to this adjusted dataset as MathBench$^*$.

We also design two custom datasets, named RandomQA and FACT, to evaluate the capabilities of the language models in data processing, numerical computation and factual retrieval.
Section \ref{sec.rcb} elaborates on the specifics of both datasets.

\noindent \textbf{SFT Settings.}
All the models in our experiments are trained with the open-source TRL library from Hugging Face \cite{vonwerra2022trl}.
The LoRA module \cite{hu2021lora} is used to perform SFT on the base model of Llama2-7B and Llama3-8B utilizing ToolBridge$^{\S}$ or ToolBridge, with a LoRA rank of 16.
Model training is conducted on 64 $\times$ MI200 64GB GPUs, with each processing a batch size of 2 (\emph{i.e.}, total batch size is 128).
AdamW is employed to optimize the parameters of LoRA, with a cosine learning rate scheduler, configuring the initial lr at 3e-5 and the total training epoch at 3.

\noindent \textbf{Inference Settings.}
The primary process of model inference is described in Algorithm \ref{alg.processst}.
Moreover, to handle potential tool call failures during inference, we propose to eliminate failed tool calls from the current output before conditioning the generation of further text.
All trained models are evaluated on 16 $\times$ MI200 64 GB GPUs, with the max new tokens set to 512 and the temperature set to zero.

\begin{table}[H]
    \centering
    \resizebox{0.7\textwidth}{!}{%
    \begin{tabular}{c|c|ccc}
        \hline
        Models  &Training Data (SFT) & GSM 8k   & GSM Plus  & MathBench$^*$ \\
         \hline
        Llama2-7B &-                      &13.6      & 8.9  & 18.0 \\
        Llama2-7B-Lora &ToolBridge$^{\S}$ &16.9   &9.9  &  19.1   \\
        Llama2-7B-Lora &ToolBridge        &18.1   &11.0  & 21.4 \\
        \hline 
        Llama3-8B &-                        & 52.3      & 36.9  & 33.0 \\
        Llama3-8B-Lora &ToolBridge$^{\S}$   & 53.4      & 37.8  & 35.2\\
        Llama3-8B-Lora &ToolBridge  & 55.8  & 40.0  & 37.4\\
        \hline 
    \end{tabular}}
    \caption{Ablation studies on GSM 8k, GSM Plus and MathBench$^*$ with (8 shots, CoT) setting. } \label{tab.s1}
    \vspace{-0.5cm}
\end{table}

\subsection{Results on Standard Benchmarks} \label{sec.rsb}

In this section, we conduct ablation studies on standard benchmark datasets, including GSM 8k, GSM Plus, MathBench and Stanford WebQA, 
where GSM 8k, GSM Plus and MathBench are primarily responsible for evaluating the capability of LLMs in numerical reasoning and computation,
and Stanford WebQA is primarily used to assess the ability of LLMs in factual retrieval.
Table \ref{tab.s1} and \ref{tab.s2} demonstrates the evaluation results.

\begin{table}[t]
    \centering
    \resizebox{0.55\textwidth}{!}{%
    \begin{tabular}{c|c|c}
        \hline
        Models             &Training Data (SFT)  & Stanford WebQA \\
         \hline
        Llama3-8B &-                        & 21.2     \\
        Llama3-8B-Lora &ToolBridge$^{\S}$   & 37.7   \\
        Llama3-8B-Lora &ToolBridge       & 39.9   \\
        \hline 
    \end{tabular}}
    \caption{Ablation studies on Stanford WebQA with zero-shot setting. } \label{tab.s2}
    \vspace{-0.5cm}
\end{table}
 
It is observed that the language models SFT on ToolBridge significantly outperform the baseline models.
For example, Llama3-8B SFT on ToolBridge brings 3.1\% and 2.2\% accuracy improvements to the base model of Llama3-8B and the Llama3-8B model SFT on ToolBridge$^{\S}$ when evaluating on GSM Plus, respectively.
These results indicate that SFT on ToolBridge can help enhance LLMs' abilities in handling numerical computations.
Furthermore, ToolBridge is able to facilitate LLMs' capabilities in factual retrieval.
As shown in Table \ref{tab.s2}, SFT on ToolBridge enables Llama3-8B to boost its performance on Stanford WebQA from 21.2\% to 39.9\%.

The results above demonstrate that LLMs can effectively learn how to utilize external tools to enhance their capabilities in basic functions after SFT on ToolBridge.
Besides, it is worthy noting that there is significant room for improvements in these results, as our focus is on the training data, with minimal adjustments made to the model structures and training strategies, 
which may help LLMs better learn how to employ external tools through ToolBridge.

\begin{table}[H]
    \centering
    \resizebox{1.0\textwidth}{!}{%
    \begin{tabular}{c|c|ccc}
        \hline
        Models  &Training Data (SFT) & RandomQA-1000-Batch1  & RandomQA-1000-Batch2 & RandomQA-1000-Batch3  \\
         \hline
        Llama2-7B &-           & 6.9  & 5.0  & 5.3 \\
        Llama2-7B-Lora &ToolBridge$^{\S}$ & 11.9    & 9.4 & 10.0\\
        Llama2-7B-Lora &ToolBridge  & \textbf{57.9}    & \textbf{55.3} & \textbf{55.9}\\
        \hline 
        Llama3-8B    &-           & 9.2      &5.9  &7.3 \\
        Llama3-8B-Lora &ToolBridge$^{\S}$ & 22.4     & 19.0  & 20.4 \\
        Llama3-8B-Lora &ToolBridge  & \textbf{69.2}   & \textbf{69.2} & \textbf{68.4} \\
        \hline 
    \end{tabular}}
    \caption{Experimental results on RandomQA under zero-shot setting.} \label{tab.randomqazeroshot}
    \vspace{-0.5cm}
\end{table}
 
\subsection{Results on Custom Benchmarks} \label{sec.rcb}

To further assess whether SFT on the ToolBridge dataset can equip LLMs with the ability to leverage external tools for aiding its reasoning process,
we propose to design two custom datasets to evaluate LLMs' performance before and after SFT on the ToolBridge dataset, namely, the RandomQA dataset and the FACT dataset.

\noindent \textbf{RandomQA.} To assess the improvements in LLMs' data processing and numerical computation capabilities after SFT on ToolBridge, 
we propose to design 50 templates capable of generating question-answer pairs to validate the abilities of the large language models in these areas. Here are some examples, 

\lstset{
  language=Python,
  basicstyle=\small\ttfamily, % Font size and typewriter font
  frame=lines, % Adds a frame around the code
  numbers=left, % Line numbers on the left
  breaklines=true, % Automatic line breaking
  keywordstyle=\color{blue}, % Color for keywords
  commentstyle=\color{gray}, % Color for comments
  stringstyle=\color{red}, % Color for strings
  showstringspaces=false,
}
\begin{lstlisting}
# Template1: Generate the smallest prime number greater than x
num = random.randint(2000, 100000)
question = f"Generate the smallest prime number greater than {num}."
answer = nextprime(num)
# Template2: Calculate time difference between two time zones
tz1, tz2 = random.sample(pytz.all_timezones, 2)
now = datetime.datetime.now()
time1 = pytz.timezone(tz1).localize(now)
time2 = pytz.timezone(tz2).localize(now)
time_difference = abs((time1 - time2).total_seconds())
question = f'Calculate time difference between {tz1} and {tz2} in seconds.'
answer = time_difference
# Template3: Find the sum of all elements above the main diagonal of a matrix
matrix_len = random.randint(2, 10)
matrix = [[random.randint(1000, 1000000) for _ in range(matrix_len)] for _ in range(matrix_len)]
question = f"Find the sum of all elements above the main diagonal of the matrix {matrix}."
answer = sum(matrix[i][j] for i in range(matrix_len) for j in range(i + 1, matrix_len))
# Template4: Sum all odd numbers in a list
array = [random.randint(1000, 1000000) for _ in range(random.randint(5, 15))]
question = f"Sum all the odd numbers in the list {array}."
answer = sum(x for x in array if x % 2 != 0)
\end{lstlisting}

The complete list can be found in Appendix \ref{sec.randomqatemplate}. 
With these pre-defined templates, we first generate three RandomQA datasets, each comprising 1,000 data entries.
For convenience, we name these three datasets RandomQA-1000-Batch1, RandomQA-1000-Batch2 and RandomQA-1000-Batch3.
Then, we evaluate the accuracy of LLMs on the three datasets before and after SFT on ToolBridge.
Table \ref{tab.randomqazeroshot} demonstrates the results.
It is observed that after SFT on ToolBridge, the models shows a significant increase in accuracy on RandomQA.

\begin{table}[H]
    \centering
    \resizebox{0.9\textwidth}{!}{%
    \begin{tabular}{c|c|ccc}
        \hline
        Models  &Training Data (SFT) & FACT-200-Batch1  & FACT-200-Batch2 & FACT-200-Batch3  \\
         \hline
        Llama2-7B &-           & 69.5   &  55.0  & 49.0 \\
        Llama2-7B-Lora &ToolBridge$^{\S}$ & 86.0    & 67.5 & 65.7 \\
        Llama2-7B-Lora &ToolBridge  & 88.5    & 72.5 & 73.2 \\
        \hline 
        Llama3-8B    &-           & 79.0      & 60.5  & 62.2 \\
        Llama3-8B-Lora &ToolBridge$^{\S}$ & 89.0   & 73.0  & 73.7 \\
        Llama3-8B-Lora &ToolBridge  & 90.0      & 73.5 & 80.2 \\
        \hline 
    \end{tabular}}
    \caption{Experimental results on FACT under zero-shot setting.} \label{tab.factzeroshot}
    \vspace{-0.5cm}
\end{table}

\noindent \textbf{FACT.} To determine if the factual retrieval skills of LLMs can be improved by SFT on ToolBridge, we propose to build the FACT dataset.
Specifically, we begin by prompting GPT-4o to produce thousands of question-answer pairs focused on factual retrieval. Some examples are as shown following,

\begin{center}
\DefineVerbatimEnvironment{BVerbatim}{Verbatim}{breaklines=true}
\begin{BVerbatim}
1. Generate 100 Q&A pairs for LLM factual retrieval testing. The question topic should be related with Geography. Return them as a Python dictionary, with concise answers (3-5 words).
2. Generate 100 Q&A pairs for LLM factual retrieval testing. The question topic should be related with History. Return them as a Python dictionary, with concise answers (3-5 words).
\end{BVerbatim}
\end{center}

Appendix \ref{sec.promptsfact} contains the entire set of the leveraged prompts for constructing FACT.
Upon obtaining these candidate question-answer pairs, we continuously draw random data entries from them and manually check their correctness until 200 correct data entries are verified.
By repeating the above process three times, we collect three FACT datasets, each with a size of 200 data entries, termed FACT-200-Batch1, FACT-200-Batch2 and FACT-200-Batch3.
Table \ref{tab.factzeroshot} provides a comparison of LLMs' performance on the three datasets before and after undergoing SFT on ToolBridge, which shows that LLMs demonstrates enhanced abilities in retrieving factual content.

Besides, one should be aware that relying on external tools to improve the fact retrieval performance of the models can be inconsistent.
During the course of our experiments, we uncover the following deficiencies,

\begin{itemize}
    \item Using external tools like the \emph{requests} module to fetch information from external websites can lead to potentially unreliable information and LLMs thus yield incorrect results. 
    A solution to address this issue is to implement a curated list of trustworthy domains, restricting LLMs from accessing websites beyond this approved whitelist.
    Additionally, further refining the training data entries to exclude unreliable access links is essential.

    \item The content scraped from web pages may be overly lengthy, hindering the ability of the large language models to efficiently extract useful information from the scraped contents.
   
    \item When LLMs access some external websites, they may attempt to download files from those sites, which poses potential security risks in real-world applications.
\end{itemize}

Therefore, we recommend that ToolBridge is only used for academic research, with controlled measures in place when it attempts to access external websites.

\begin{figure}[t]
\centering
\includegraphics[width=1.0\textwidth]{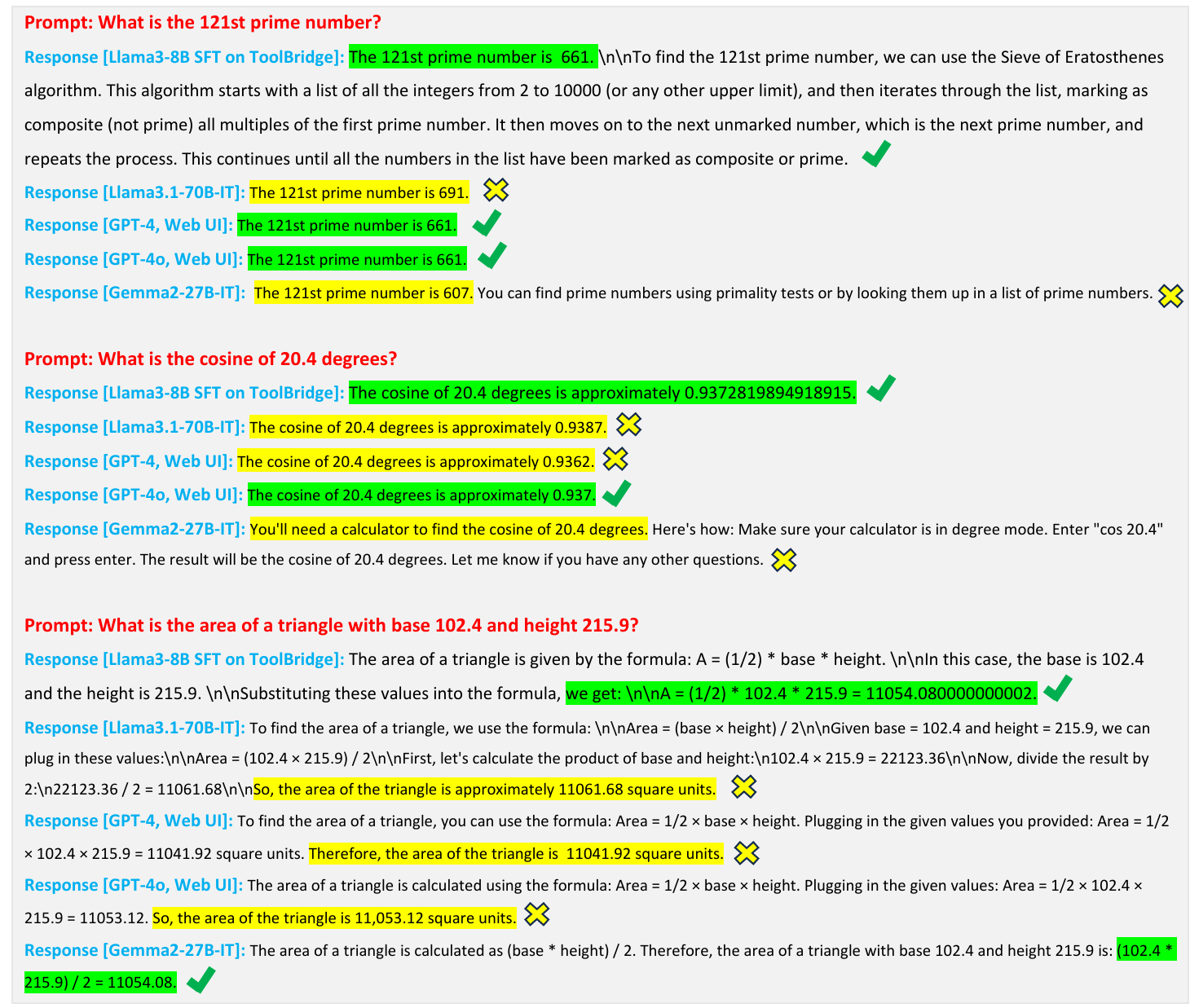} 
\vspace{-0.5cm}
\caption{\textbf{A qualitative comparison with other language models.} IT denotes instruct and Web UI means we directly prompt the model using the official user interface provided in the browser.}
\label{fig.outputexamples}
\end{figure}

\subsection{Qualitative results} \label{sec.qr}

Finally, we provide a comparison of the qualitative results from Llama3-8B against other LLMs, both those capable of utilizing external tools to aid its reasoning process (\emph{e.g.}, Llama3.1 and GPT-4o) and those that are not (\emph{e.g.}, Gemma2).

As shown in Figure \ref{fig.outputexamples}, we can observe that, after SFT on ToolBridge, Llama3-8B demonstrates significant advantages in areas such as numerical computation, owing to its ability to utilize external tools to facilitate its reasoning process.
For example, when calculating trigonometric functions, Llama3-8B SFT on ToolBridge can obtain accurate trigonometric values for each angle by invoking the Python \emph{math} library, 
while LLMs that are unable to invoke external tools can only provide an answer with significant errors.

Moreover, some additional quantitative comparison results can be found in Appendix \ref{sec.morequaltitativeexamples}.

\section{Conclusion}

In this paper, we seek to shed light on the process of constructing datasets for training LLMs in the use of external tools.
Specifically, we propose a pipeline to build the ToolBridge dataset, where the foundation of ToolBridge lies in the raw data entries for SFT, sourced from the community.
The selection, conversion, and filtering of these raw data entries are crucial steps in developing high-quality training data for equipping LLMs with external tool abilities.
Our experiments demonstrate that SFT on ToolBridge enables LLMs to effectively learn how to utilize external tools to aid in reasoning process, particularly in the areas including data processing, numerical computation and factual retrieval.
We will make the data generation pipeline for training LLMs in tool use publicly available, along with ToolBridge, for the benefit of the community in advancing research on enabling LLMs to learn how to use external tools.

%Bibliography
\newpage
\bibliographystyle{unsrt}  
\bibliography{references}

\newpage
\appendix

\section{Appendix}
\DefineVerbatimEnvironment{BVerbatim}{Verbatim}{breaklines=true}

\subsection{Llama3-70B Prompts} \label{sec.llama3prompts}

The prompt applied while using Llama3-70B to identify valuable data is as follow,

\begingroup
\fontsize{9pt}{10pt}\selectfont
\begin{BVerbatim}
Your task is to determine whether you can add calls to a Python API to a piece of text. The calls should help you get information required to complete the text. You only need to respond with "Yes" or "No", "Yes" means you can and "No" means you can't. Here are some examples:

Input: 

{"messages": [{"role": "user", "content": "Sort the numbers in the list arr = [1, 10, 2, 5, -2, 11, 12] in descending order."}, {"role": "assistant", "content": "The sorted list in descending order is: [12, 11, 10, 5, 2, 1, -2]."}]}

Output:

Yes

Input: 

{"messages": [{"role": "user", "content": "Can you tell me a little bit about what LaTeX is?"}, {"role": "assistant", "content": "LaTeX is a high-quality typesetting system; it includes features designed for the production of technical and scientific documentation. LaTeX is the de facto standard for the communication and publication of scientific documents. It is widely used by mathematicians, scientists, engineers, philosophers, linguists, economists, and other scholars in academia and the professional world."}]}

Output:

No

Input: 

{"messages": [{"role": "user", "content": "What is the value of sin 40 degrees plus cos 31 degrees?"}, {"role": "assistant", "content": "The value is approximately sin 40 + cos 31 = 0.6428 + 0.8572 = 1.500."}]}

Output:

Yes

Input:

{"messages": [{"role": "user", "content": "Write a Python script that reads an image and recognizes the text on it."}, {"role": "assistant", "content": "To read an image and recognize text on it in Python, you can use the pytesseract library along with Pillow for image processing. Here's a simple example: \nimport pytesseract; from PIL import Image; print(pytesseract.image_to_string(Image.open('path_to_image.jpg'))). \nMake sure you have Tesseract OCR installed on your machine and the required libraries (pytesseract and Pillow) installed in your Python environment. You can install them using: \npip install pytesseract pillow"}]}

Output:

No

Input:

PLACEHOLDER

Output:
\end{BVerbatim}
\endgroup

\newpage

\subsection{GPT-4o-mini Prompts} \label{sec.gpt40miniprompts}

\begingroup

The prompt applied while using GPT-4o-mini to convert data entries is as follow,

\fontsize{8pt}{9pt}\selectfont
\begin{BVerbatim}
Your task is to add calls to a Python API to a piece of text. The calls should help you get information required to complete the text. You can call the API by writing "<python>code</python>" where "code" is the code to be executed. The last line of all code should print the variable that stores the final result. Here are some examples of API calls:

Input: 

{"messages": [{"role": "user", "content": "Which number is greater, 13.11 or 13.8?"}, {"role": "assistant", "content": "13.8 is greater than 13.11."}]}

Output: 

{"messages": [{"role": "user", "content": "Which number is greater, 13.11 or 13.8?"}, {"role": "assistant", "content": "<python>greater_number = max(13.11, 13.8)\nprint(greater_number)</python> 13.8 is greater than 13.11."}]}

Input: 

{"messages": [{"role": "user", "content": "How many unique words are there in the sentence 'The quick brown fox jumps over the lazy dog'?"}, {"role": "assistant", "content": "There are eight unique words in the sentence 'The quick brown fox jumps over the lazy dog.'"}]}

Output:

{"messages": [{"role": "user", "content": "How many unique words are there in the sentence 'The quick brown fox jumps over the lazy dog'?"}, {"role": "assistant", "content": "There are <python>unique_words = len(set('The quick brown fox jumps over the lazy dog'.lower().split()))\nprint(unique_words)</python> eight unique words in the sentence 'The quick brown fox jumps over the lazy dog.'"}]}

Input:

{"messages": [{"role": "user", "content": "What is the area of a circle with a radius of 5?"}, {"role": "assistant", "content": "The area of a circle with radius 5 is 78.54."}]}

Output:

{"messages": [{"role": "user", "content": "What is the area of a circle with a radius of 5?"}, {"role": "assistant", "content": "The area of a circle with radius 5 is <python>import math\narea = math.pi * 5**2\nprint(area)</python> 78.54."}]}

Input:

{"messages": [{"role": "user", "content": "Sort the numbers [5, 3, 8, 1, 2] in ascending order."}, {"role": "assistant", "content": "The sorted list is [1, 2, 3, 5, 8]."}]}

Output:

{"messages": [{"role": "user", "content": "Sort the numbers [5, 3, 8, 1, 2] in ascending order."}, {"role": "assistant", "content": "The sorted list is <python>lst = sorted([5, 3, 8, 1, 2])\nprint(lst)</python> [1, 2, 3, 5, 8]."}]}

Input:

{"messages": [{"role": "user", "content": "Extract the domain from the email 'example@test.com'."}, {"role": "assistant", "content": "The domain of the email 'example@test.com' is 'test.com'."}]}

Output:

{"messages": [{"role": "user", "content": "Extract the domain from the email 'example@test.com'."}, {"role": "assistant", "content": "The domain of the email 'example@test.com' is <python>domain = 'example@test.com'.split('@')[1]\nprint(domain)</python> 'test.com'."}]}

Input:

PLACEHOLDER

Output:
\end{BVerbatim}
\endgroup

\newpage

\subsection{Detecting Simple Code Structures: Assignment Followed by Print Statement} \label{sec.identifyingcodepatterns}

We adopt the following algorithm to determine if the inserted code in each data entry is simply an assignment followed by printing the variable, which is meaningless.

\begin{algorithm}
\caption{Identifying Code Patterns: Distinguishing Assignments Followed by Print Statements} 
\label{alg.identifyingcodepatterns}
\begin{lstlisting}[language=Python, basicstyle=\ttfamily\footnotesize, 
    keywordstyle=\color{blue}\bfseries, stringstyle=\color{red}, commentstyle=\color{green!50!black}, 
    numbers=left, numberstyle=\tiny\color{gray}, 
    xleftmargin=2em, xrightmargin=2em, breaklines=true, tabsize=4]
import ast

'''
Function:
    Identifying Code Patterns: Distinguishing Assignments Followed by Print Statements
Argument:
    node: node = ast.parse(code)
'''
def isuselesscode(node):
    # Check if the node is an AST Module
    if isinstance(node, ast.Module):
        # Ensure the module has exactly two statements:
        # an assignment and an expression.
        if len(node.body) == 2 and isinstance(node.body[0], ast.Assign) and isinstance(node.body[1], ast.Expr):
            assign_node = node.body[0]
            expr_node = node.body[1]
            # Check if the assignment targets a variable and 
            # the value is a constant.
            if isinstance(assign_node.targets[0], ast.Name) and isinstance(assign_node.value, ast.Constant):
                # Check if the expression is a function call 
                # to 'print'.
                if isinstance(expr_node.value, ast.Call) and isinstance(expr_node.value.func, ast.Name) and expr_node.value.func.id == 'print':
                    # Ensure 'print' has exactly one argument.
                    if len(expr_node.value.args) == 1:
                        arg = expr_node.value.args[0]
                        # Check if the argument to 'print' is 
                        # the same variable assigned earlier.
                        if isinstance(arg, ast.Name) and arg.id == assign_node.targets[0].id:
                            return True
                        # Alternatively, check if 'print' uses an 
                        # f-string format with the variable.
                        elif isinstance(arg, ast.JoinedStr):
                            for value in arg.values:
                                if isinstance(value, ast.FormattedValue) and isinstance(value.value, ast.Name):
                                    # Confirm the formatted  
                                    # variable is the same as 
                                    # the assigned variable.
                                    if value.value.id == assign_node.targets[0].id:
                                        return True
    return False
\end{lstlisting}
\end{algorithm}

\newpage

\subsection{Optimizing Dataset Quality by Removing Non-Executable and Slow Tool Calls} \label{sec.removebadcode}

The core code for removing data entries that either fail to execute or exceed the execution time limit is as follows.

\begin{algorithm}
\caption{Efficient Dataset Curation: Filtering Non-Executable and Time-Consuming Tool Calls} 
\label{alg.removebadcode}
\begin{lstlisting}[language=Python, basicstyle=\ttfamily\footnotesize, 
    keywordstyle=\color{blue}\bfseries, stringstyle=\color{red}, commentstyle=\color{green!50!black}, 
    numbers=left, numberstyle=\tiny\color{gray}, 
    xleftmargin=2em, xrightmargin=2em, breaklines=true, tabsize=4]
import re
import io
import contextlib
import multiprocessing

# Execute Python code
def executecode(code):
    with io.StringIO() as buf, contextlib.redirect_stdout(buf):
        try:
            exec(code)
            return buf.getvalue().strip()
        except Exception:
            return None

# Execute Python code with timeout = 30 second
def safeexecutecode(code, timeout=30):
    result_queue = multiprocessing.Queue()
    def target():
        result = executecode(code)
        result_queue.put(result)
    process = multiprocessing.Process(target=target)
    process.start()
    process.join(timeout)
    if process.is_alive():
        process.terminate()
        process.join()
        return None
    else:
        return result_queue.get() if not result_queue.empty() else None

# Filtering
DATASET_FILTERED = []
for item in tqdm(DATASET):
    success_calling = False
    for message in item['messages']:
        if message['role'] == 'assistant':
            code_matches = re.findall(r'<python>(.*?)</python>', message['content'], re.DOTALL)
            if code_matches:
                for code in code_matches:
                    python_str = f'<python>{code}</python>'
                    result = safeexecutecode(code=code)
                    if result is None:
                        message['content'] = message['content'].replace(python_str, '')
                    else:
                        result_str = f'<result>{result}</result>'
                        message['content'] = message['content'].replace(python_str, f'{python_str}{result_str}')
                        success_calling = True
    if success_calling:
        DATASET_FILTERED.append(item)
\end{lstlisting}
\end{algorithm}

\newpage
\subsection{Compare Outputs from Llama3-8B Trained on $\mathcal{C}$ and ToolBridge}\label{sec.compareconsistentoutputs}

We notice in actual applications that the models SFT on $\mathcal{C}$ may not always leverage the tool execution results as a basis for generating further content.
Here are some examples.

\begin{figure}[h]
\centering
\includegraphics[width=1.0\textwidth]{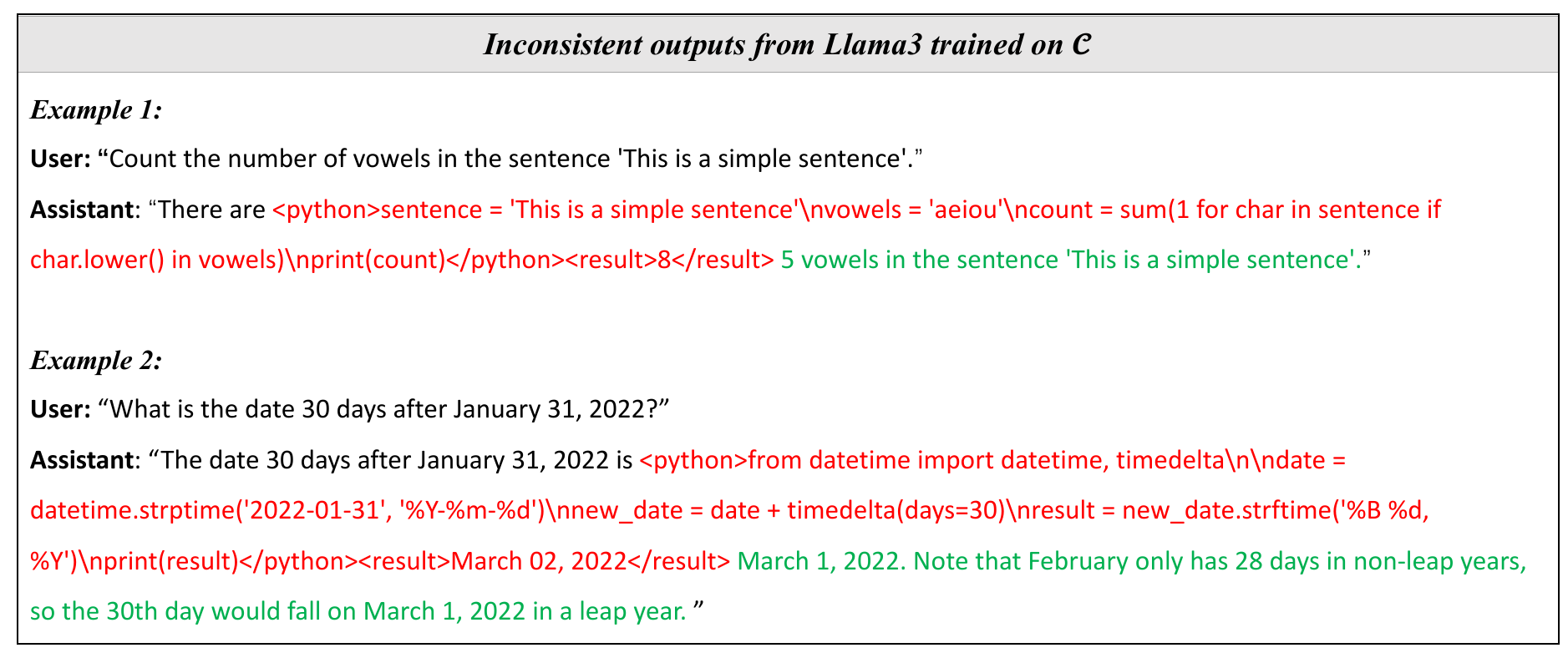} \label{fig.inconsistencyoutputs}
\caption{Inconsistency between LLMs' outputs and the returned results from Python interpreter. }
\end{figure}

By filtering out the data entries in $\mathcal{C}$ that show inconsistencies between the Python execution results and the following content, we create the ToolBridge dataset.
This refined dataset is then leveraged to train the Llama3-8B model, yielding outputs consistent with the tool execution results.
Here are some examples.

\begin{figure}[h]
\centering
\includegraphics[width=1.0\textwidth]{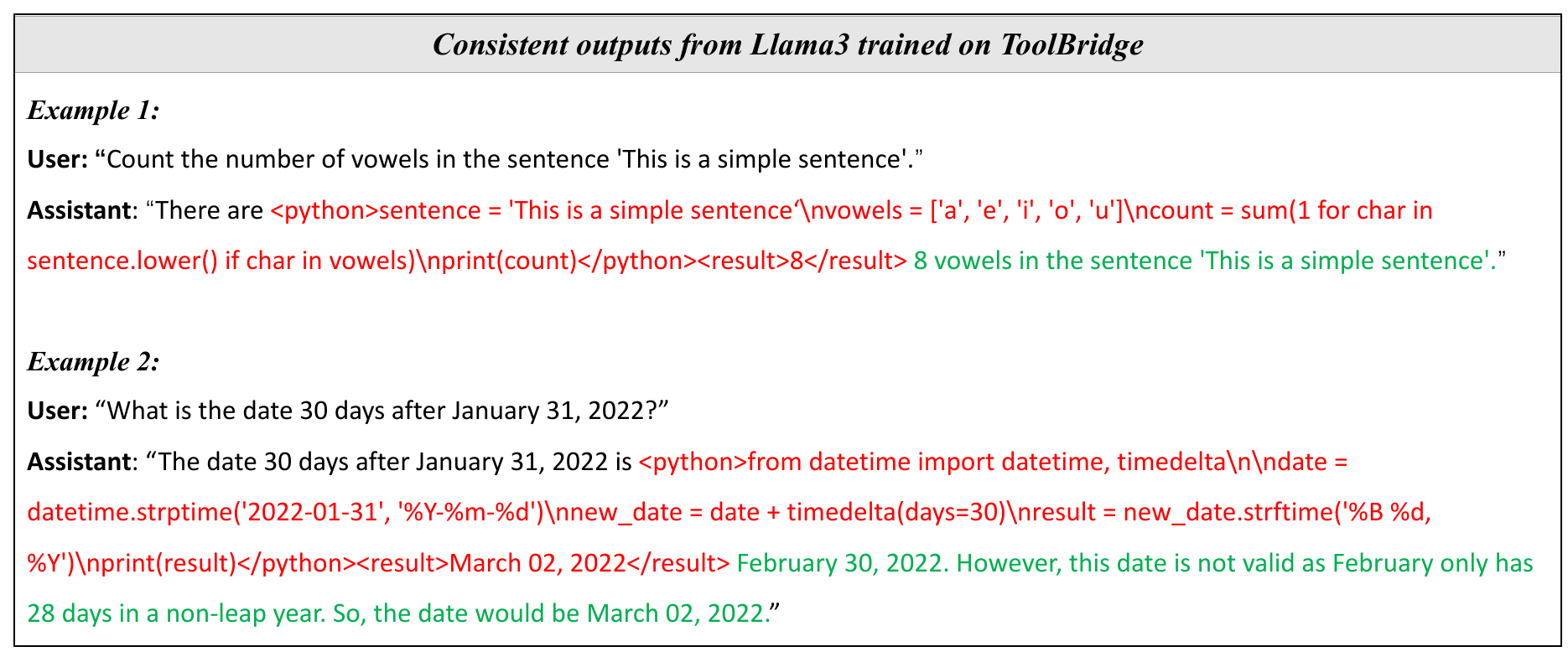} \label{fig.consistencyoutputs}
\caption{Consistency between LLMs' outputs and the returned results from Python interpreter. }
\end{figure}

\newpage
\subsection{Links to Data Sources Used in Our Dataset Pool}\label{sec.datasourcelink}

To avoid any confusion regarding the adopted datasets in this study, the download links for each dataset we have used are provided in Table \ref{tab.datasourcelink}. 
Note that, links to all datasets in the dataset pool $\mathcal{P}$ are presented in this table, even those not incorporated into ToolBridge in the end.

\begin{table}[h]
    \centering
    \resizebox{1.0\textwidth}{!}{%
    \begin{tabular}{c|c}
        \hline
        \textbf{Dataset} & \textbf{Source Link} \\ \hline
        School Math 0.25M  & \texttt{https://huggingface.co/datasets/BelleGroup/school\_math\_0.25M} \\ \hline
        ChatAlpaca        & \texttt{https://github.com/cascip/ChatAlpaca} \\ \hline
        Platypus         & \texttt{https://huggingface.co/datasets/garage-bAInd/Open-Platypus} \\ \hline  
        ShareGPT90K      & \texttt{https://huggingface.co/datasets/liyucheng/ShareGPT90K} \\ \hline 
        WizardLM\_Orca   & \texttt{https://huggingface.co/datasets/pankajmathur/WizardLM\_Orca} \\ \hline 
        WizardLM\_evol\_instruct\_70k   &\texttt{https://huggingface.co/datasets/WizardLMTeam/WizardLM\_evol\_instruct\_70k} \\ \hline 
        WizardLM evolve\_instruct V2   &\texttt{https://huggingface.co/datasets/WizardLMTeam/WizardLM\_evol\_instruct\_V2\_196k} \\ \hline 
        OpenOrca &\texttt{https://huggingface.co/datasets/Open-Orca/OpenOrca} \\ \hline 
        TigerBot &\texttt{https://huggingface.co/datasets/TigerResearch/sft\_en}, \texttt{TigerResearch/sft\_zh}  \\ \hline 
        GPT-4all &\texttt{https://huggingface.co/datasets/nomic-ai/gpt4all-j-prompt-generations} \\ \hline 
        COIG &\texttt{https://huggingface.co/datasets/BAAI/COIG} \\ \hline 
        LIMA &\texttt{https://huggingface.co/datasets/GAIR/lima} \\ \hline 
        AlpacaDataCleaned &\texttt{https://huggingface.co/datasets/yahma/alpaca-cleaned} \\ \hline 
        GPT-4-LLM &\texttt{https://github.com/Instruction-Tuning-with-GPT-4/GPT-4-LLM} \\ \hline 
        Bactrian-X &\texttt{https://huggingface.co/datasets/MBZUAI/Bactrian-X} \\ \hline 
        code\_instructions\_120k\_alpaca &\texttt{https://huggingface.co/datasets/iamtarun/code\_instructions\_120k\_alpaca} \\ \hline 
        TSI-v0    &\texttt{https://huggingface.co/datasets/tasksource/tasksource-instruct-v0} \\ \hline 
        Alpaca  &\texttt{https://github.com/tatsu-lab/stanford\_alpaca} \\ \hline 
        No Robots     &\texttt{https://huggingface.co/datasets/HuggingFaceH4/no\_robots} \\ \hline 
        Baize    &\texttt{https://github.com/project-baize/baize-chatbot} \\ \hline 
        LaMini-Instruction    &\texttt{https://huggingface.co/datasets/MBZUAI/LaMini-instruction} \\ \hline 
        tiny-codes     &\texttt{https://huggingface.co/datasets/nampdn-ai/tiny-codes} \\ \hline 
        self-instruct    &\texttt{https://github.com/yizhongw/self-instruct} \\ \hline 
        ign\_clean\_instruct\_dataset\_500k    &\texttt{https://huggingface.co/datasets/ignmilton/ign\_clean\_instruct\_dataset\_500k} \\ \hline 
        MOSS SFT    &\texttt{https://github.com/OpenMOSS/MOSS} \\ \hline 
    \end{tabular}}
    \caption{Source links for the datasets utilized in the dataset pool.}
    \label{tab.datasourcelink}
\end{table}

\newpage

\subsection{Templates for Yielding RandomQA Dataset}\label{sec.randomqatemplate}

The complete list of templates used to generate the RandomQA dataset is as follows,

\begin{lstlisting}
'''RandomQAGenerator'''
class RandomQAGenerator():
    question_types = [
        "Calculate the average of an array", 
        "Find the maximum and minimum values of an array", 
        "Calculate the dot product of two arrays", 
        "Sort an array in ascending order",
        "Generate a set of random integers and find their sum", 
        "Generate the smallest prime number greater than x", 
        "Calculate the standard deviation of a list of floating-point numbers",
        "Generate a random matrix and find its inverse",
        "Count the frequency of one character in a string",
        "Square every number in a list",
        "Find the median of an array",
        "Generate Fibonacci sequence up to n-th term",
        "Transpose a 2D matrix",
        "Reverse the string",
        "Find the GCD (Greatest Common Divisor) of two numbers",
        "Calculate the factorial of a number",
        "Find the mode of a list of numbers",
        "Calculate the sum of even numbers in a list",
        "Calculate the cumulative sum of an array",
        "Extract first N elements in a list",
        "Calculate cosine value",
        "Reverse the order of elements in a list",
        "Calculate the sum of squares of numbers in an array",
        "Find the n-th smallest number in an array",
        "Calculate the Euclidean distance between two points in a plane",
        "Find the intersection of two strings",
        "Calculate the compound interest given principal, rate, and time",
        "Find the length of the longest word in a string",
        "Count the number of vowels in a string",
        "Convert a list of Celsius temperatures to Fahrenheit",
        "Calculate time difference between two time zones",
        "Find the leap year after a year",
        "Find the most common word in a paragraph",
        "Calculate the perimeter of a rectangle given its length and width",
        "Sum all the digits of a given number",
        "Calculate the area of a triangle given its base and height",
        "Find the real roots of a quadratic equation",
        "Calculate the sum of the cubes of a list",
        "Round all elements in a list to two decimal places",
        "Find the first recurring word in a string",
        "Calculate the hypotenuse of a right triangle given the other two sides",
        "Extract all the numbers in a string",
        "Convert a decimal number to its binary equivalent",
        "Calculate the difference between two lists",
        "Sum all odd numbers in a list",
        "Find out all the numbers that are not unique",
        "Flatten a 2D list into a 1D list",
        "Remove duplicates from a list",
        "Generate the smallest N primes",
        "Find the sum of all elements above the main diagonal of a matrix"
    ]
    def __init__(self, num_gen_qa=1000):
        self.num_gen_qa = num_gen_qa
    '''generate'''
    def generate(self):
        qa_pairs = []
        for _ in range(self.num_gen_qa):
            qa_pairs.append(self.randomgenone())
        pickle.dump(qa_pairs, open(f'random_qa_{int(time.time())}.pkl', 'wb'))
    '''randomgenone'''
    def randomgenone(self, question_type=None):
        # randomly choose a question type if question_type is None
        question_type = random.choice(self.question_types) if question_type is None else question_type
        # generate question and answer based on type
        # 1. Calculate the average of an array
        if question_type == "Calculate the average of an array":
            array = [round(random.uniform(-10000, 10000)) for _ in range(random.randint(5, 15))]
            question = f"Calculate the average of the array {array} and round the result to two decimal places."
            answer = round(sum(array) / len(array), 2)
        # 2. Find the maximum and minimum values of an array
        elif question_type == "Find the maximum and minimum values of an array":
            array = [round(random.uniform(-10000, 10000)) for _ in range(random.randint(5, 15))]
            max_or_min = random.choice(['maximum', 'minimum'])
            question = f"Find the {max_or_min} value of the array {array}, give the result of multiplying it by 7."
            answer = max(array) if max_or_min == 'maximum' else min(array)
            answer = answer * 7
        # 3. Calculate the dot product of two arrays
        elif question_type == "Calculate the dot product of two arrays":
            length = random.randint(5, 15)
            array1 = [random.randint(20, 1000) for _ in range(length)]
            array2 = [random.randint(20, 1000) for _ in range(length)]
            question = f"Calculate the dot product of the arrays {array1} and {array2}."
            answer = sum(x * y for x, y in zip(array1, array2))
        # 4. Sort an array in ascending order
        elif question_type == "Sort an array in ascending order":
            array = [random.randint(-10000, 10000) for _ in range(random.randint(5, 15))]
            question = f"Sort the array {array} in ascending order."
            answer = sorted(array)
        # 5. Generate a set of random integers and find their sum
        elif question_type == "Generate a set of random integers and find their sum":
            array = [random.randint(1000, 100000) for _ in range(random.randint(5, 15))]
            question = f"Here is a set of random integers {array}, please find their sum."
            answer = sum(array)
        # 6. Generate the smallest prime number greater than x
        elif question_type == "Generate the smallest prime number greater than x":
            num = random.randint(2000, 100000)
            question = f"Generate the smallest prime number greater than {num}."
            answer = nextprime(num)
        # 7. Calculate the standard deviation of a list of floating-point numbers
        elif question_type == "Calculate the standard deviation of a list of floating-point numbers":
            array = [round(random.uniform(10, 1000), 2) for _ in range(random.randint(5, 15))]
            mean = sum(array) / len(array)
            variance = sum((x - mean) ** 2 for x in array) / len(array)
            question = f"Calculate the standard deviation of the array {array} and round the result to two decimal places."
            answer = round(variance ** 0.5, 2)
        # 8. Generate a random matrix and find its inverse
        elif question_type == "Generate a random matrix and find its inverse":
            matrix_len = random.randint(2, 10)
            matrix = [[random.randint(1, 1000) for _ in range(matrix_len)] for _ in range(matrix_len)]
            question = f"Here is a random matrix {matrix}, please find its inverse, you can answer with 'not invertible' if its inverse does not exist."
            det = np.linalg.det(matrix)
            if int(det) != 0:
                inv_matrix = np.linalg.inv(matrix).tolist()
            else:
                inv_matrix = "not invertible"
            answer = inv_matrix
        # 9. Count the frequency of one character in a string
        elif question_type == "Count the frequency of one character in a string":
            char = random.choice('abcdefghijklmnopqrstuvwxyz')
            string = ''.join(random.choices('abcdefghijklmnopqrstuvwxyz', k=random.randint(50, 100))) + char * 101
            question = f"Count the frequency of character {char} in the string '{string}'."
            answer = string.count(char)
        # 10. Square every number in a list
        elif question_type == "Square every number in a list":
            array = [random.randint(1, 10000) for _ in range(random.randint(5, 15))]
            question = f"Square every number in the list {array}."
            answer = [x ** 2 for x in array]
        # 11. Find the median of an array
        elif question_type == "Find the median of an array":
            array = [random.randint(200000, 10000000) for _ in range(random.randint(5, 15))]
            sorted_array = sorted(array)
            question = f"Find the median of the array {array}, give the result of multiplying it by 9."
            answer = sorted_array[len(sorted_array) // 2]
            answer = answer * 9
        # 12. Generate Fibonacci sequence up to n-th term
        elif question_type == "Generate Fibonacci sequence up to n-th term":
            n = random.randint(5, 20)
            question = f"Generate the Fibonacci sequence up to the {n}-th term."
            fib = [0, 1]
            for i in range(2, n):
                fib.append(fib[-1] + fib[-2])
            answer = fib
        # 13. Transpose a 2D matrix
        elif question_type == "Transpose a 2D matrix":
            matrix_len = random.randint(2, 10)
            matrix = [[random.randint(-1000, 1000) for _ in range(matrix_len)] for _ in range(matrix_len)]
            question = f"Transpose the matrix {matrix}."
            answer = [list(row) for row in zip(*matrix)]
        # 14. Reverse the string
        elif question_type == "Reverse the string":
            string = ''.join(random.choices('abcdefghijklmnopqrstuvwxyz', k=random.randint(10, 20)))
            question = f"Reverse the string {string}, and splice it behind the string 'appleiphone'."
            answer = 'appleiphone' + string[::-1]
        # 15. Find the GCD (Greatest Common Divisor) of two numbers
        elif question_type == "Find the GCD (Greatest Common Divisor) of two numbers":
            answer = 0
            while answer <= 100:
                a, b = random.randint(200, 1000000), random.randint(200, 1000000)
                question = f"Find the GCD of the numbers {a} and {b}."
                answer = math.gcd(a, b)
        # 16. Calculate the factorial of a number
        elif question_type == "Calculate the factorial of a number":
            num = random.randint(10, 100)
            question = f"Calculate the factorial of {num}."
            answer = math.factorial(num)
        # 17. Find the mode of a list of numbers
        elif question_type == "Find the mode of a list of numbers":
            array = [random.randint(113333, 113343) for _ in range(15)]
            question = f"Find the mode of the array {array}, give the result of multiplying it by 3."
            answer = max(set(array), key=array.count)
            answer = answer * 3
        # 18. Calculate the sum of even numbers in a list
        elif question_type == "Calculate the sum of even numbers in a list":
            array = [random.randint(1000, 1000000) for _ in range(random.randint(10, 25))]
            question = f"Calculate the sum of even numbers in the list {array}."
            answer = sum(x for x in array if x % 2 == 0)
        # 19. Calculate the cumulative sum of an array
        elif question_type == "Calculate the cumulative sum of an array":
            array = [random.randint(1, 10000) for _ in range(random.randint(5, 15))]
            question = f"Calculate the cumulative sum of the array {array}."
            answer = [sum(array[:i+1]) for i in range(len(array))]
        # 20. Extract first N elements in a list
        elif question_type == "Extract first N elements in a list":
            N = random.randint(5, 10)
            array = [random.randint(1, 10000) for _ in range(random.randint(15, 35))]
            question = f"Extract first {N} elements in the list {array} and then plus 7 for each element in the sub-list."
            answer = array[:N]
            answer = [a + 7 for a in answer]
        # 21. Calculate cosine value
        elif question_type == "Calculate cosine value":
            degree = random.randint(0, 360) + 0.5
            question = f"Calculate cosine value for {degree} degree and round the result to two decimal places."
            answer = round(math.cos(math.radians(degree)), 2)
        # 22. Reverse the order of elements in a list
        elif question_type == "Reverse the order of elements in a list":
            array = [random.randint(1, 10000) for _ in range(random.randint(5, 15))]
            question = f"Reverse the order of the elements in the list {array} and then plus 3 for each element."
            answer = array[::-1]
            answer = [a + 3 for a in answer]
        # 23. Calculate the sum of squares of numbers in an array
        elif question_type == "Calculate the sum of squares of numbers in an array":
            array = [random.randint(10, 10000) for _ in range(random.randint(5, 15))]
            question = f"Calculate the sum of squares of the numbers in the array {array}."
            answer = sum(x ** 2 for x in array)
        # 24. Find the n-th smallest number in an array
        elif question_type == "Find the n-th smallest number in an array":
            array = [random.randint(1000, 10000000) for _ in range(random.randint(5, 15))]
            n = random.randint(1, len(array))
            question = f"Find the {n}-th smallest number in the array {array}, give the result of multiplying it by 3."
            answer = sorted(array)[n - 1] * 3
        # 25. Calculate the Euclidean distance between two points in a plane
        elif question_type == "Calculate the Euclidean distance between two points in a plane":
            x1, y1 = round(random.uniform(-100, 100), 2), round(random.uniform(-100, 100), 2)
            x2, y2 = round(random.uniform(-100, 100), 2), round(random.uniform(-100, 100), 2)
            question = f"Calculate the Euclidean distance between points ({x1}, {y1}) and ({x2}, {y2}), round the result to two decimal places."
            answer = round(math.sqrt((x2 - x1)**2 + (y2 - y1)**2), 2)
        # 26. Find the intersection of two strings
        elif question_type == "Find the intersection of two strings":
            str1 = ''.join(random.choices('abcdefghijklmnopqrstuvwxyz', k=random.randint(50, 100)))
            str2 = ''.join(random.choices('abcdefghijklmnopqrstuvwxyz', k=random.randint(50, 100)))
            question = f"Find the intersection of string '{str1}' and string '{str2}'."
            answer = ''.join(set(str1) & set(str2))
        # 27. Calculate the compound interest given principal, rate, and time
        elif question_type == "Calculate the compound interest given principal, rate, and time":
            principal = random.randint(1000, 10000)
            rate = round(random.uniform(1, 10), 2)
            time = random.randint(1, 5)
            question = f"Calculate the compound interest for principal {principal}, rate {rate}%, and time {time} years, round the result to two decimal places."
            answer = round(principal * (1 + rate/100)**time, 2)
        # 28. Find the length of the longest word in a string
        elif question_type == "Find the length of the longest word in a string":
            words = [''.join(random.choices('abcdefghijklmnopqrstuvwxyz', k=random.randint(101, 200))) for _ in range(random.randint(5, 15))]
            string = ' '.join(words)
            question = f"Find the length of the longest word in the string '{string}'."
            answer = max(len(word) for word in words)
        # 29. Count the number of vowels in a string
        elif question_type == "Count the number of vowels in a string":
            string = ''.join(random.choices('abcdefghijklmnopqrstuvwxyz', k=random.randint(20, 50))) + 'a' * 101
            question = f"Count the number of vowels in the string '{string}'."
            answer = sum(1 for char in string if char in 'aeiou')
        # 30. Convert a list of Celsius temperatures to Fahrenheit
        elif question_type == "Convert a list of Celsius temperatures to Fahrenheit":
            celsius_list = [random.randint(-20, 40) for _ in range(5)]
            question = f"Convert the list of Celsius temperatures {celsius_list} to Fahrenheit, round the result to two decimal places."
            answer = [round(c * 9/5 + 32, 2) for c in celsius_list]
        # 31. Calculate time difference between two time zones
        elif question_type == "Calculate time difference between two time zones":
            tz1, tz2 = random.sample(pytz.all_timezones, 2)
            now = datetime.datetime.now()
            time1 = pytz.timezone(tz1).localize(now)
            time2 = pytz.timezone(tz2).localize(now)
            time_difference = abs((time1 - time2).total_seconds())
            question = f'Calculate time difference between {tz1} and {tz2} in seconds.'
            answer = time_difference
        # 32. Find the leap year after a year
        elif question_type == "Find the leap year after a year":
            year = random.randint(1900, 2100)
            while calendar.isleap(year):
                year = random.randint(1900, 2100)
            question = f"Find the leap year after year {year}."
            answer = next(y for y in range(year + 1, year + 10000) if calendar.isleap(y))
        # 33. Find the most common word in a paragraph
        elif question_type == "Find the most common word in a paragraph":
            words = ['apple', 'banana', 'orange', 'grape', 'pear', 'hello', 'iphone', 'newspaper']
            paragraph = ' '.join(random.choices(words, k=30))
            question = f"Find the most common word in the paragraph '{paragraph}', concatenate it with the second common word in this paragraph."
            answer = Counter(paragraph.lower().split()).most_common(2)[0][0] + Counter(paragraph.lower().split()).most_common(2)[1][0]
        # 34. Calculate the perimeter of a rectangle given its length and width
        elif question_type == "Calculate the perimeter of a rectangle given its length and width":
            length, width = random.randint(100, 10000), random.randint(100, 10000)
            question = f"Calculate the perimeter of a rectangle with length {length} and width {width}."
            answer = 2 * (length + width)
        # 35. Sum all the digits of a given number
        elif question_type == "Sum all the digits of a given number":
            num = int(str(random.randint(100, 99999)) + '999999999999999')
            question = f"Sum all the digits of the number {num}."
            answer = sum(int(digit) for digit in str(num))
        # 36. Calculate the area of a triangle given its base and height
        elif question_type == "Calculate the area of a triangle given its base and height":
            base = round(random.uniform(100, 500), 2)
            height = round(random.uniform(100, 500), 2)
            question = f"Calculate the area of a triangle with base {base} and height {height}, round the result to two decimal places."
            answer = round(0.5 * base * height, 2)
        # 37. Find the real roots of a quadratic equation
        elif question_type == "Find the real roots of a quadratic equation":
            a = round(random.uniform(10, 200), 2)
            b = round(random.uniform(10, 200), 2)
            c = round(random.uniform(10, 200), 2)
            question = f"Find the real roots of the quadratic equation {a}x^2 + {b}x + {c} = 0, round the result to two decimal places."
            discriminant = b**2 - 4*a*c
            if discriminant > 0:
                root1 = (-b + math.sqrt(discriminant)) / (2*a)
                root2 = (-b - math.sqrt(discriminant)) / (2*a)
                answer = (round(root1, 2), round(root2, 2))
            elif discriminant == 0:
                root = -b / (2*a)
                answer = round(root, 2)
            else:
                answer = "no real roots"
        # 38. Calculate the sum of the cubes of a list
        elif question_type == "Calculate the sum of the cubes of a list":
            sequence = [random.randint(100, 10000) for _ in range(random.randint(5, 15))]
            question = f"Calculate the sum of the cubes of the list {sequence}."
            answer = sum([n**3 for n in sequence])
        # 39. Round all elements in a list to two decimal places
        elif question_type == "Round all elements in a list to two decimal places":
            array = [random.uniform(100, 10000) for _ in range(random.randint(5, 15))]
            question = f"Round all elements in the list {array} to two decimal places."
            answer = [round(num, 2) for num in array]
        # 40. Find the first recurring word in a string
        elif question_type == "Find the first recurring word in a string":
            words = [''.join(random.choices('abcdefghijklmnopqrstuvwxyz', k=random.randint(5, 15))) for _ in range(random.randint(5, 10))]
            words = words * 3
            random.shuffle(words)
            paragraph = ' '.join(words)
            question = f"Find the first recurring word in the paragraph '{paragraph}', concatenate it with the second recurring word in this paragraph."
            def _find_recurring_words(paragraph):
                words = paragraph.lower().split()
                seen = set()
                first, second = None, None
                for word in words:
                    if word in seen:
                        if first is None:
                            first = word
                        elif second is None and word != first:
                            second = word
                            break
                    seen.add(word)
                return first + second
            answer = _find_recurring_words(paragraph)
        # 41. Calculate the hypotenuse of a right triangle given the other two sides
        elif question_type == "Calculate the hypotenuse of a right triangle given the other two sides":
            side1 = random.randint(100, 20000)
            side2 = random.randint(100, 20000)
            question = f"Calculate the hypotenuse of a right triangle with sides {side1} and {side2}, round the result to two decimal places."
            answer = round(math.sqrt(side1**2 + side2**2), 2)
        # 42. Extract all the numbers in a string
        elif question_type == "Extract all the numbers in a string":
            string1 = random.choices('abcdefghijklmnopqrstuvwxyz', k=random.randint(20, 50))
            string2 = random.choices('0123456789', k=random.randint(20, 50))
            string = string1 + string2
            random.shuffle(string)
            string = ''.join(string)
            question = f"Extract all the numbers in the string '{string}' in order and concatenate them."
            answer = ''.join(re.findall(r'\d+', string))
        # 43. Convert a decimal number to its binary equivalent
        elif question_type == "Convert a decimal number to its binary equivalent":
            num = random.randint(1000, 1000000)
            question = f"Convert the decimal number {num} to its binary equivalent."
            answer = bin(num)[2:]
        # 44. Calculate the difference between two lists
        elif question_type == "Calculate the difference between two lists":
            list1 = [random.randint(1, 50) for _ in range(10)]
            list2 = [random.randint(1, 50) for _ in range(10)]
            question = f"Calculate the difference between the lists {list1} and {list2}."
            answer = list(set(list1) - set(list2))
        # 45. Sum all odd numbers in a list
        elif question_type == "Sum all odd numbers in a list":
            array = [random.randint(1000, 1000000) for _ in range(random.randint(5, 15))]
            question = f"Sum all the odd numbers in the list {array}."
            answer = sum(x for x in array if x % 2 != 0)
        # 46. Find out all the numbers that are not unique
        elif question_type == "Find out all the numbers that are not unique":
            array = [random.randint(20, 35) for _ in range(20)]
            question = f"Find out all the numbers that are not unique in the array {array}."
            answer = [num for num, count in Counter(array).items() if count > 1]
        # 47. Flatten a 2D list into a 1D list
        elif question_type == "Flatten a 2D list into a 1D list":
            array_len = random.randint(2, 10)
            array = [[random.randint(1, 1000) for _ in range(array_len)] for _ in range(array_len)]
            question = f"Flatten the 2D list {array} into a 1D list."
            answer = [item for sublist in array for item in sublist]
        # 48. Remove duplicates from a list
        elif question_type == "Remove duplicates from a list":
            array = [random.randint(1, 20) for _ in range(15)]
            while len(array) == len(set(array)):
                array = [random.randint(1, 20) for _ in range(15)]
            question = f"Remove duplicates from the list {array}."
            answer = list(set(array))
        # 49. Generate the smallest N primes
        elif question_type == "Generate the smallest N primes":
            n = random.randint(5, 20)
            primes = []
            candidate = 2
            while len(primes) < n:
                if all(candidate % i != 0 for i in range(2, int(candidate ** 0.5) + 1)):
                    primes.append(candidate)
                candidate += 1
            question = f"Generate the smallest {n} prime numbers."
            answer = primes
        # 50. Find the sum of all elements above the main diagonal of a matrix
        elif question_type == "Find the sum of all elements above the main diagonal of a matrix":
            matrix_len = random.randint(2, 10)
            matrix = [[random.randint(1000, 1000000) for _ in range(matrix_len)] for _ in range(matrix_len)]
            question = f"Find the sum of all elements above the main diagonal of the matrix {matrix}."
            answer = sum(matrix[i][j] for i in range(matrix_len) for j in range(i + 1, matrix_len))
        # not defined question 
        else:
            raise ValueError(f'{question_type} is not defined')
        # format and return
        random_qa = {'question': question, 'answer': answer}
        return random_qa
\end{lstlisting}

\newpage
\subsection{Prompts for Construing FACT Dataset} \label{sec.promptsfact}

We construct the FACT datasets by prompting GPT-4o with the following instructions,

\begin{center}
\DefineVerbatimEnvironment{BVerbatim}{Verbatim}{breaklines=true}
\begin{BVerbatim}
1. Generate 100 Q&A pairs for LLM factual retrieval testing. The question topic should be related with Geography. Return them as a Python dictionary, with concise answers (3-5 words).
2. Generate 100 Q&A pairs for LLM factual retrieval testing. The question topic should be related with History. Return them as a Python dictionary, with concise answers (3-5 words).
3. Generate 100 Q&A pairs for LLM factual retrieval testing. The question topic should be related with Science. Return them as a Python dictionary, with concise answers (3-5 words).
4. Generate 100 Q&A pairs for LLM factual retrieval testing. The question topic should be related with Technology. Return them as a Python dictionary, with concise answers (3-5 words).
5. Generate 100 Q&A pairs for LLM factual retrieval testing. The question topic should be related with Mathematics. Return them as a Python dictionary, with concise answers (3-5 words).
6. Generate 100 Q&A pairs for LLM factual retrieval testing. The question topic should be related with Culture and Arts. Return them as a Python dictionary, with concise answers (3-5 words).
7. Generate 100 Q&A pairs for LLM factual retrieval testing. The question topic should be related with Sports. Return them as a Python dictionary, with concise answers (3-5 words).
8. Generate 100 Q&A pairs for LLM factual retrieval testing. The question topic should be related with Politics. Return them as a Python dictionary, with concise answers (3-5 words).
9. Generate 100 Q&A pairs for LLM factual retrieval testing. The question topic should be related with Language and Grammar. Return them as a Python dictionary, with concise answers (3-5 words).
10. Generate 100 Q&A pairs for LLM factual retrieval testing. The question topic should be related with Current Affairs. Return them as a Python dictionary, with concise answers (3-5 words).
11. Generate 100 Q&A pairs for LLM factual retrieval testing. The question topic should be related with Entertainment. Return them as a Python dictionary, with concise answers (3-5 words).
12. Generate 100 Q&A pairs for LLM factual retrieval testing. The question topic should be related with Medicine and Health. Return them as a Python dictionary, with concise answers (3-5 words).
13. Generate 100 Q&A pairs for LLM factual retrieval testing. The question topic should be related with Economics and Business. Return them as a Python dictionary, with concise answers (3-5 words).
14. Generate 100 Q&A pairs for LLM factual retrieval testing. The question topic should be related with Religion and Mythology. Return them as a Python dictionary, with concise answers (3-5 words).
15. Generate 100 Q&A pairs for LLM factual retrieval testing. The question topic should be related with General Knowledge. Return them as a Python dictionary, with concise answers (3-5 words).
\end{BVerbatim}
\end{center}

\newpage

\subsection{Qualitative Comparison with Other LLMs} \label{sec.morequaltitativeexamples}

Here, we show more qualitative comparison between Llama3-8B SFT on ToolBridge and GPT-4o and Gemini.

\begin{figure}[h]
\centering
\includegraphics[width=1.0\textwidth]{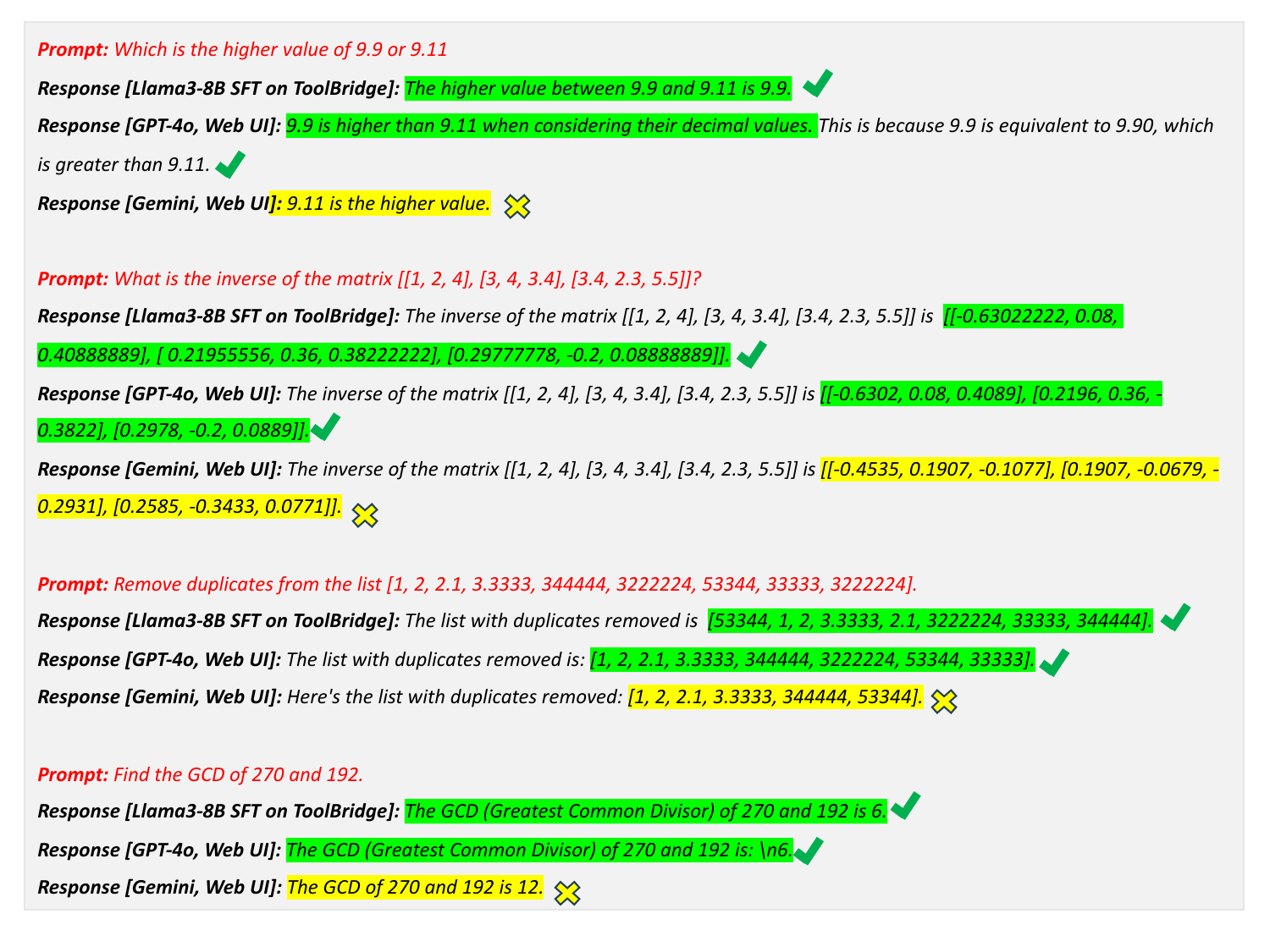} \label{fig.moreexamples}
\caption{Qualitative comparison with GPT-4o and Gemini. }
\end{figure}

\end{document}